\documentclass[conference]{IEEEtran}
\usepackage[noadjust]{cite}
\usepackage{nopageno}
\usepackage[T1]{fontenc}
\usepackage[pdftex]{graphicx}
\graphicspath{{./figures/}}
\usepackage{balance}
\usepackage{footnote}
\usepackage{amsmath}
\usepackage{booktabs}
\usepackage{url}
\usepackage{balance}
\usepackage[cmintegrals]{newtxmath}
\begin{document}
	\title{Key Ingredients of Self-Driving Cars}
	\author{
		Rui Fan, Jianhao Jiao, Haoyang Ye, Yang Yu, Ioannis Pitas, Ming Liu
		\thanks{R. Fan, J. Jiao, H. Ye, Y. Yu and M. Liu are with the Robotics and Multi-Perception Laboratory in Robotics Institute at the Hong Kong University of Science and Technology, Hong Kong. }
		\thanks{I. Pitas is with the Department of Informatics at Aristotle University of Thessaloniki, Thessaloniki, Greece. }
	}
	\maketitle
\begin{abstract}
Over the past decade, many research articles have been published in the area of autonomous driving. However, most of them focus only on a specific technological area, such as visual environment perception, vehicle control, etc. 
Furthermore, due to fast advances in the self-driving car technology, such articles become obsolete very fast. In this paper, we give  a brief but comprehensive overview on key ingredients of autonomous cars (ACs), including driving automation levels, AC sensors, AC software, open source datasets, industry leaders,  AC applications and existing challenges. 
\end{abstract}
\section{Introduction}
\label{sec.introduction}
Over the past decade, with a number of autonomous system technology breakthroughs being witnessed in the world, the race to commercialize Autonomous Cars (ACs) has become fiercer than ever \cite{Brink2017}. For example, in 2016, Waymo unveiled its autonomous taxi service in Arizona, which has attracted large publicity \cite{waymo2018}. 
Furthermore, Waymo has spent around nine years in developing and improving its Automated Driving Systems (ADSs) using various advanced engineering technologies, e.g., machine learning and computer vision \cite{waymo2018}.
These cutting-edge technologies have greatly assisted their driver-less vehicles in better world understanding, making the right decisions, and taking the right actions at the right time \cite{waymo2018}. 
	
\begin{figure}[t!]
	\centering
	\includegraphics[width=0.47\textwidth]{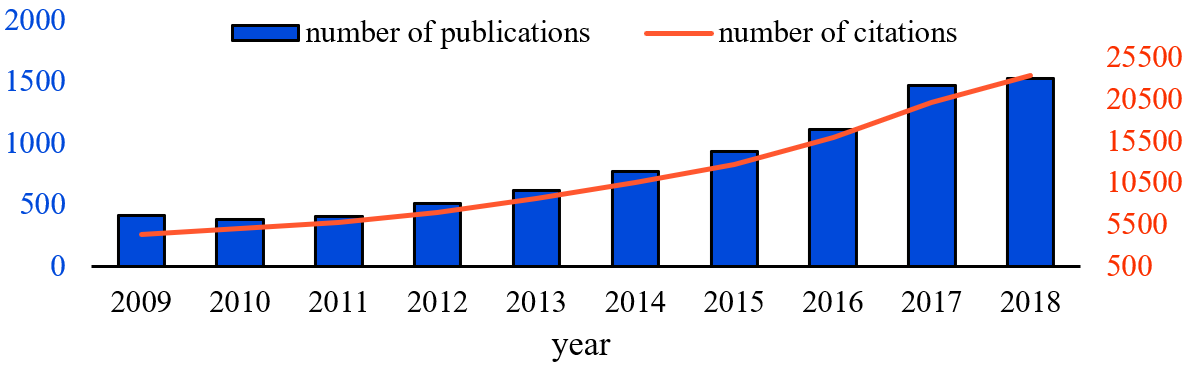}
	\caption{Numbers of publications and citations in autonomous driving research over the past decade.}
	\vspace{-1.5em}
	\label{fig:year_pub_citation}
\end{figure}

Owing to the development of autonomous driving, many scientific articles have been published over the past decade, and their citations\footnote{\url{https://www.webofknowledge.com}} are increasing exponentially, as shown in Fig. \ref{fig:year_pub_citation}.  We can clearly see that the numbers of both publications and citations in each year have been increasing gradually since 2010 and rose to a new height in the last year.  However, most of autonomous driving overview articles focus only on a specific technological area, such as  Advanced Driver Assistance Systems (ADAS) \cite{okuda2014survey}, vehicle control \cite{gonzalez2016review}, visual environment perception \cite{mukhtar2015vehicle}, etc. Therefore, there is a strong motivation to provide readers with a comprehensive literature review on autonomous driving, including  systems and algorithms, open source datasets, industry leaders, autonomous car applications and existing challenges.
\section{AC Systems}
\label{sec.systems_algorithms}
ADSs enable ACs to operate in a real-world environment without intervention by Human Drivers (HDs). Each ADS consists of two main components: hardware (car sensors and hardware controllers, i.e., throttle, brake, wheel, etc.) and software (functional groups).  
	
\begin{figure}[t!]
	\centering\includegraphics[width=0.48\textwidth]{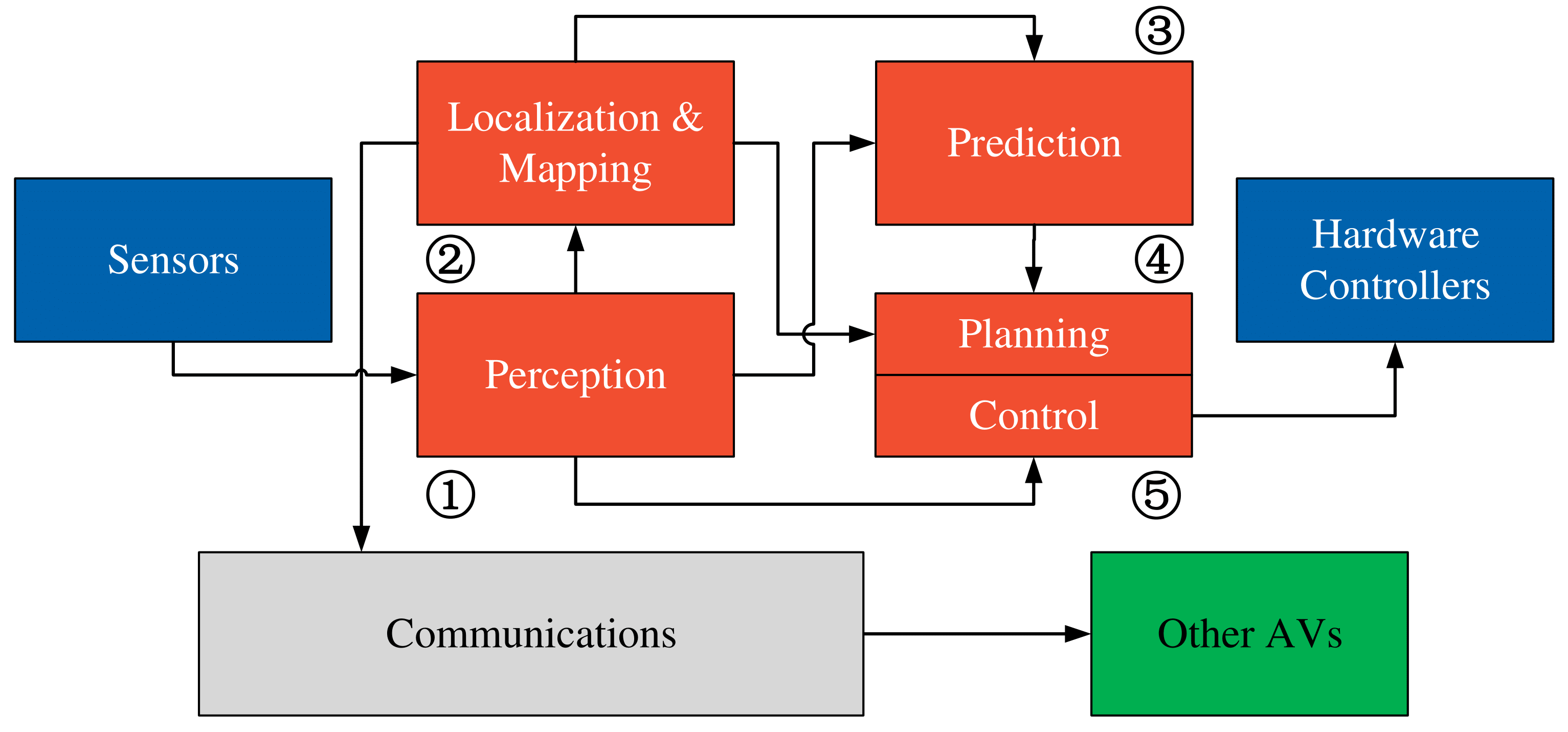}\caption{Software architecture of our proposed ADS.}
	\label{fig.ads}
	\vspace{-1.5em}
\end{figure}

Software has been modeled in several different software architectures, such as Stanley (Grand Challenge) \cite{thrun2006stanley}, Junior (Urban Challenge) \cite{montemerlo2008junior}, Boss (Urban Challenge) \cite{urmson2008autonomous} and Tongji AC \cite{zong2018architecture}. Stanley \cite{thrun2006stanley} software architecture   comprises four modules: sensor interface, perception, planning and control, as well as user interface. Junior \cite{montemerlo2008junior} software architecture has five parts: sensor interface, perception, navigation (planning and control), drive-by-wire interface (user interface and vehicle interface) and global services. Boss \cite{urmson2008autonomous} uses a three-layer architecture: mission, behavioral and motion planning. Tongji's ADS \cite{zong2018architecture} partitions the software architecture in: perception, decision and planning, control and chassis. In this paper, we divide the software architecture into  five modules: perception, localization and mapping, prediction, planning and control, as shown in Fig. \ref{fig.ads}, which is very similar to Tongji ADS's software architecture \cite{zong2018architecture}. The remainder of this section introduces  driving automation levels, presents the AC sensors and hardware controllers.

\subsection{Driving Automation Levels}
\label{sec.driving_automation_level}
	
According to the Society of Automotive Engineers (SAE international), driving automation can be categorized into six levels, as shown in Table \ref{tab.sae_level} \cite{sae2016taxonomy}. 
HD is responsible for Driving Environment Monitoring (DEM) in level 0-2 ADSs, while this responsibility shifts to the system in level 3-5 ADSs.  From level 4, the HD is not responsible for the Dynamic Driving Task Fallback (DDTF) any more, and  the ADSs will not need to ask for intervention from the HD at level 5. The state-of-the-art ADSs are mainly at level 2 and 3. A long time may still be needed to achieve higher automation levels \cite{hecht2018}. 
	
\begin{table}[!b]
			\vspace{-2em}
\caption{SAE Levels of Driving Automation}
\begin{tabular}{lllll}
			\hline
			\toprule
			Level & Name                   & Driver           & DEM     & DDTF     \\ \hline \toprule
			0     & No automation          & HD           & HD     & HD     \\ 
			1     & Driver assistance      & HD \& system & HD     & HD     \\ 
			2     & Partial automation     & System       & HD     & HD     \\ 
			3     & Conditional automation & System       & System & HD     \\ 
			4     & High automation        & System       & System & System \\ 
			5     & Full automation        & System       & System & System \\ 
			\hline
			\toprule
		\end{tabular}

		\label{tab.sae_level}
		\centering
	\end{table}
	
\subsection{AC Sensors}
\label{sec.sensors}
The sensors mounted on ACs are generally used to perceive the environment. Each sensor is chosen as a trade-off between sampling rate, Field of View (FoV), accuracy, range, cost and overall system complexity \cite{sensor_set}. The most commonly used AC sensors are passive ones (e.g., cameras), active ones (e.g., Lidar, Radar and ultrasonic transceivers) and other sensor types, e.g., Global Positioning System (GPS), Inertial Measurement Unit (IMU) and wheel encoders \cite{sensor_set}. 

Cameras capture 2D images by collecting light reflected on the 3D environment objects. The image quality  is usually subject to the environmental conditions, e.g., weather and illumination.  Computer vision and machine learning algorithms are generally used to extract useful information from captured images/videos \cite{Fan2018}. For example, the images captured from different view points, i.e., using a single movable camera or multiple synchronized cameras, can be used to acquire 3D world geometry information \cite{Fan2018e}. 

Lidar illuminates a target with pulsed laser light and measures the source distance to the target, by analyzing the reflected pulses \cite{lidar}. Due to its high 3D geometry accuracy, Lidar is generally used to create high-definition world maps \cite{lidar}. Lidars are usually mounted on different parts of the AC, e.g., roof, side and front, for different purposes \cite{Jiao2019a, jiao2019automatic}. Radars can measure accurate range and radial velocity of an object, by transmitting an electromagnetic wave and analyzing the reflected one \cite{Bureau2013}. They are particularly good at detecting metallic objects, but can also detect non-metallic objects, such as pedestrians and trees, in a short distance \cite{sensor_set}.  Radars have been established in the automotive industry for many years to enable ADAS features, such as autonomous emergency braking, adaptive cruise control, etc \cite{sensor_set}.  In a similar way to Radar, ultrasonic transducers calculate the distance to an object by measuring the time between transmitting an ultrasonic signal and receiving its echo \cite{Westerveld2014}. Ultrasonic transducers are commonly utilized for AC localization and navigation \cite{Liu2018}.  
	
GPS, a satellite-based radio-navigation system owned by the US government, can provide time and geolocation information for ACs. However, GPS signals are very weak and they can be easily blocked by obstacles, such as buildings and mountains, resulting in GPS-denied regions, e.g., in the so-called urban canyons \cite{Samama2008}. Therefore, IMUs are commonly integrated into GPS devices to ensure AC localization in such places \cite{CoxJr1978}. Wheel encoders are also prevalently utilized to determine the AC position, speed and direction by measuring  electronic signals regarding wheel motion \cite{Trahey2008}. 
\subsection{Hardware Controllers}
\label{sec.hardware_controllers}
AC hardware controllers are torque steering motor, electronic brake booster, electronic throttle, gear shifter and parking brake. The vehicle states, such as wheel speed and steering angle, are sensed automatically and sent to the computer system via a Controller Area Network (CAN) bus. This enables either the HD or the ADS to control  throttle, brake and steering wheel \cite{Bhandari2010}.
\section{AC Software}
\subsection{Perception}
\label{sec.perception}
The perception module analyzes the raw sensor data and outputs an environment understanding to be used by the ACs \cite{Maurer2016}.  This process is similar to human visual cognition. Perception module typically includes object (free space, lane, vehicle, pedestrian, road damage, etc) detection and tracking, 3D world reconstruction (using structure from motion, stereo vision, etc), among others \cite{Fan2016, Fan2017}. The state-of-the-art perception technologies can be broken into two  categories: computer vision-based and machine learning-based ones \cite{Apolloni2005}. The former generally addresses visual perception problems by formulating them with explicit projective geometry models and finding the best solution using optimization approaches. For example, in \cite{Ozgunalp2017}, the horizontal and vertical coordinates of multiple vanishing points were modeled using a parabola and a quartic polynomial, respectively. 
The lanes were then  detected using these two polynomial functions. Machine learning-based technologies learn the best solution to a given perception problem, by employing  data-driven classification and/or regression models, such as the Convolutional Neural Networks (CNNs) \cite{Chen2015}. For instance, some deep CNNs, e.g., SegNet \cite{Badrinarayanan2017} and U-Net \cite{Ronneberger2015}, have achieved impressive performance in semantic image segmentation and object classification. Such CNNs can also be easily utilized for other similar perception tasks using transfer learning (TL) \cite{Maurer2016}. Visual world perception can be complemented by using other sensors, e.g., Lidars or Radars, for obstacle detection/localization and for 3D world modeling. Multi-sensor information fusion for world perception can produce superior world understanding results. 
\subsection{Localization and Mapping}
\label{sec.loc_map}
Using sensor data and perception output, the localization and mapping module can not only estimate AC location, but also build and update a 3D world map \cite{Maurer2016}. This topic has become very popular since the concept of Simultaneous Localization and Mapping (SLAM) was introduced in 1986 \cite{Smith1986}. The state-of-the-art SLAM systems are generally classified as filter-based \cite{ye2019tightly} and optimization-based \cite{bresson2017simultaneous}. The filter-based SLAM systems are derived from Bayesian filtering \cite{bresson2017simultaneous}.  They iteratively estimate AC pose and update the 3D environmental map,  by incrementally integrating sensor data. The most commonly used filters are Extended Kalman Filter (EKF) \cite{Kalman1960}, Unscented Kalman Filter (UKF) \cite{Julier1997}, Information Filter (IF) \cite{Maybeck1980} and Particle Filter (PF) \cite{Dellaert1999}. On the other hand, the optimization-based SLAM approaches firstly identify the problem constraints by finding a correspondence between new observations and the map. Then, they compute and refine AC previous poses and update the 3D environmental map. The optimization-based SLAM approaches can be  divided into two main branches: Bundle Adjustment (BA) and graph SLAM \cite{bresson2017simultaneous}. The former one jointly optimizes the 3D world map and the camera poses by minimizing an error function using optimization techniques, such as the Gaussian-Newton method and Gradient Descent \cite{Rao2009}. The latter one models the localization problem as a graph representation problem and solves it by finding an error function with respect to different vehicle poses \cite{Thrun2008}. 
\subsection{Prediction}
\label{sec.prediction}
The prediction module analyzes the motion patterns of other traffic agents and predicts AC future trajectories \cite{ma2018trafficpredict}, which enables the AC to make appropriate navigation decisions. Current prediction approaches can be grouped into two main categories: model-based and data-driven-based \cite{lefevre2014survey}. The former computes the AC future motion, by propagating its kinematic state (position, speed and acceleration) over time, based on the underlying physical system kinematics and dynamics \cite{lefevre2014survey}. For example, Mercedes-Benz motion prediction component employs map information as a constraint to compute the next AC position \cite{bojarski2016end}. A Kalman filter \cite{kalman1960new} works well for short-term predictions, but its performance degrades for long-term horizons, as it ignores surrounding context, e.g., roads and traffic rules \cite{djuric2018motion}.
Furthermore,  a pedestrian motion prediction model can be formed based on attractive and repulsive forces \cite{helbing1995social}. With  recent advances in Artificial Intelligence (AI) and High-Performance Computing (HPC), many data-driven techniques, e.g., the Hidden Markov Models (HMM) \cite{streubel2014prediction}, Bayesian Networks (BNs) \cite{schreier2016integrated} and Gaussian Process (GP) regression, have been utilized to predict AC states. In recent years, researchers have modeled the environmental context using Inverse Reinforcement Learning (IRL) \cite{ng2000algorithms}.  For example,  an inverse optimal control method was employed in \cite{kitani2012activity} to predict pedestrian paths.

\subsection{Planning}
\label{sec.planning}
The planning module determines possible safe AC navigation routes based on perception, localization and mapping, as well as prediction information \cite{Katrakazas2015}. The planning tasks can mainly be classified as path, maneuver and trajectory \cite{Paden2016}. Path is a list of geometrical way points  that the AC should follow, so as to reach its destination, without colliding with obstacles \cite{Gonzalez2016}. The most commonly used path planning techniques include: Dijkstra \cite{Cormen2001}, dynamic programming \cite{Jiao2019}, A* \cite{Delling2009}, state lattice \cite{Gonzalez-Sieira2014}, etc. Maneuver planning is a high-level AC motion characterization process, because it also takes traffic rules and other AC states into consideration \cite{Gonzalez2016}. The trajectory is generally represented by a sequence of AC states. A trajectory satisfying the motion model and state constraints must be generated after finding the best path and maneuver, because this can ensure traffic safety and comfort.

\subsection{Control}
\label{sec.control}
The control module sends appropriate commands to throttle, braking, or steering torque, based on the predicted trajectory  and the estimated vehicle states \cite{Gruyer2016}. The control module enables the AC to follow the planned trajectory as closely as possible. The controller parameters can be estimated by minimizing an error function (deviation) between the ideal and observed states. The most prevalently used approaches to minimize such error function are Proportional-Integral-Derivative (PID) control \cite{Araki2009}, Linear-Quadratic Regulator (LQR) control \cite{Goodwin2001} and Model
Predictive Control (MPC) \cite{Garcia1989}. A PID controller is a control loop feedback mechanism, which employs proportional, integral and derivative terms to minimize the error function \cite{Araki2009}. LQR controller is utilized to minimize the error function, when the system dynamics are represented by a set of linear differential equations and the cost is described by a quadratic function \cite{Goodwin2001}. MPC is an advanced process control technique which relies on a dynamic process model \cite{Garcia1989}. These three controllers have their own benefits and drawbacks. AC control module generally employs a mixture of them. For example, Junior AC \cite{levinson2011towards} employs  MPC and PID to complete some low-level feedback control tasks, e.g., for applying the torque converter to achieve a desired wheel angle. Baidu Apollo employs a mixture of these three controllers:  PID is used for feed-forward control; LQR is used for wheel angle control; MPC is used to optimize PID and LQR controller parameters \cite{Huang2018}.

\section{Open Source Datasets}
\label{sec.datasets}
Over the past decade, many open source datasets have been  published to contribute to autonomous driving research. In this paper, we only enumerate the most cited ones. Cityscapes \cite{cordts2016cityscapes} contains a large-scale dataset which can be used for both pixel-level and instance-level semantic image segmentation. ApolloScape \cite{Huang2018}  can be used for various AC perception tasks, such as scene parsing, car instance understanding, lane segmentation, self-localization, trajectory estimation, as well as object detection and tracking. Furthermore, KITTI \cite{geiger2013vision} offers visual datasets for stereo and flow estimation, object detection and tracking, road segmentation, odometry estimation and semantic image segmentation. 6D-vision \cite{HernanBadino2011} uses a stereo camera to perceive the 3D environment. They offer datasets for stereo, optical flow and semantic image segmentation. 

\section{Industry Leaders}
\label{sec.leaders}
Recently, investors have started to throw their money at possible winners of the race to commercialize ACs \cite{news2}. Tesla's valuation has been soaring since 2016. This leads underwriters to speculate that this company will spawn a self-driving fleet in a couple of years' time \cite{news2}. In addition, GM's shares have risen by 20 percent, since their plan to build driver-less vehicles was reported  in 2017 \cite{news2}.  Waymo has tested its self-driving cars over a distance of eight million miles in the US until July 2018 \cite{news1}. Their Chrysler Pacifica mini-vans can now navigate on highways in San Francisco at full speed \cite{news2}.  GM and Waymo had the fewest accidents in the last year: GM had 22 collisions over 212 kilometers, while Waymo had only three collisions over more than 563 kilometers \cite{news1}. 
In addition to the industry giants, world-class universities have also accelerated the development of autonomous driving.  These universities are all doing well in carrying out their education with the combination of production and scientific research. This renders them better contribute to enterprises, economy and society. 
\section{AC Applications}
\label{sec.automotive_applications}
The autonomous driving technology can be implemented in any types of vehicles, such as taxis, coaches, tour buses, delivery vans, etc. These vehicles can not only relieve humans from labor-intensive and tedious work, but also ensure their safety. For example, the road quality assessment vehicles equipped with autonomous driving technology can repair the detected road damages while navigating across the city \cite{Fan2018, Fan2018c, Fan2018a, Fan2019}. Furthermore, 
public traffic will be more efficient and secure, as the coaches and taxis will be able to communicate with each other intelligently. 
\section{Existing Challenges}
\label{sec.key_challenges}
Although the autonomous driving technology has  developed rapidly over the past decade, there are still many challenges. For example, the perception modules cannot perform well in poor weather and/or illumination conditions or in complex urban environments \cite{VanBrummelen2018}. In addition, most  perception methods are generally computationally-intensive and cannot run in real time on embedded and resource-limited hardware. Furthermore, the use of current SLAM approaches still remains limited in large-scale experiments, due to its long-term unstability \cite{bresson2017simultaneous}. Another important issue is how to fuse AC sensor data to create more accurate semantic 3D word in a fast and cheap way. Moreover, ``when can people truly accept autonomous driving and autonomous cars?'' is still a good topic for discussion and poses serious ethical issues.  
\section{Conclusion}
\label{sec.conclusion}
This paper presents a brief but comprehensive overview on the key ingredients of autonomous cars. We introduced six driving automation levels. The details on the sensors equipped in autonomous cars for data collection and the hardware controllers were subsequently given.
Furthermore, we briefly discussed each software part of the ADS, i.e., perception, localization and mapping, prediction, planning, and control. The open source datasets, such as KITTI, ApolloSpace and 6D-vision, were then introduced. Finally, we discussed current autonomous driving industry leaders, the possible applications of autonomous driving and the existing challenges in this research area. \vspace{-0.3em}

\bibliographystyle{IEEEtran}

\balance	
\end{document}